\documentclass[lettersize,journal]{IEEEtran}
\usepackage{amsmath,amsfonts}
\usepackage{algorithmic}
\usepackage{array}
\usepackage[caption=false,font=normalsize,labelfont=sf,textfont=sf]{subfig}
\usepackage{textcomp}
\usepackage{stfloats}
\usepackage{url}
\usepackage{verbatim}
\usepackage{graphicx}
\usepackage{booktabs}	
\usepackage{tabularx}	
\usepackage{makecell}	
\usepackage{cite}		
\def\BibTeX{{\rm B\kern-.05em{\sc i\kern-.025em b}\kern-.08em
    T\kern-.1667em\lower.7ex\hbox{E}\kern-.125emX}}
\usepackage{balance}
\begin{document}
\title{Chaos in Motion: Unveiling Robustness in Remote Heart Rate Measurement through Brain-Inspired Skin Tracking}
\author{Jie Wang, Jing Lian, Minjie Ma, Junqiang Lei, Chunbiao Li, Bin Li, Jizhao Liu
	\thanks{This work is supported by the Natural Science Foundation of Gansu Province (Grants 21JR7RA510, 21JR7RA345, 22JR5RA543). \textit{(Corresponding author: Jizhao Liu.)} }
	\thanks{Jie Wang, Bin Li, Jizhao Liu are with the School of Information Science and Engineering, Lanzhou University, Lanzhou 730000, China (e-mail: binli@lzu.edu.cn; liujz@lzu.edu.cn). }
	\thanks{Jing Lian is with the School of Electronics and Information Engineering, Lanzhou Jiaotong University, Lanzhou 730070, China (e-mail: lian322scc@163.com).}
	\thanks{Minjie Ma is with the Department of Thoracic Surgery, The First Hospital of Lanzhou University, Lanzhou 730000, China (e-mail: maminjie24@sina.com).}
	\thanks{Junqiang Lei is with the Department of Radiology, The First Hospital of Lanzhou University, Lanzhou 730000, China (e-mail: leijq@lzu.edu.cn).}
	\thanks{Chunbiao Li is with the School of Artificial Intelligence, Nanjing University of Information Science and Technology, Nanjing 210044, China (e-mail: chunbiaolee@nuist.edu.cn).}}

\markboth{Journal of \LaTeX\ Class Files,~Vol.~18, No.~9, September~2020}%
{How to Use the IEEEtran \LaTeX \ Templates}

\maketitle

\begin{abstract}
Heart rate is an important physiological indicator of human health status. Existing remote heart rate measurement methods typically involve facial detection followed by signal extraction from the region of interest (ROI). These SOTA methods have three serious problems: (a) inaccuracies even failures in detection caused by environmental influences or subject movement; (b) failures for special patients such as infants and burn victims; (c) privacy leakage issues resulting from collecting face video. To address these issues, we regard the remote heart rate measurement as the process of analyzing the spatiotemporal characteristics of the optical flow signal in the video. We apply chaos theory to computer vision tasks for the first time, thus designing a brain-inspired framework. Firstly, using an artificial primary visual cortex model to extract the skin in the videos, and then calculate heart rate by time-frequency analysis on all pixels. Our method achieves Robust Skin Tracking for Heart Rate measurement, called HR-RST. The experimental results show that HR-RST overcomes the difficulty of environmental influences and effectively tracks the subject movement. Moreover, the method could extend to other body parts. Consequently, the method can be applied to special patients and effectively protect individual privacy, offering an innovative solution.
\end{abstract}


\section{Introduction}
\IEEEPARstart{H}{eart} rate (HR) stands as a crucial and convenient physiological indicator, with its precise measurement proving important for diagnosing cardiovascular diseases and providing daily medical care. The method of remote heart rate measurement using regular cameras offers non-contact advantages in comparison to traditional measurement techniques. This method can present innovative avenues for comprehensive cardiovascular health assessment and diagnosis, it operates on the fundamental principle that the pulsation of the heart leads to pulsations within the blood flow. As blood flows through the skin, it induces subtle, periodic color variations in the skin area. From this phenomenon, the corresponding blood volume pulse (BVP) signal \cite{ref2} can be reconstructed, facilitating the feasibility of remote heart rate measurement. However, BVP signals are exceptionally weak and susceptible to noise interference arising from fluctuations in lighting conditions and subject movements. In response to these challenges, a myriad of techniques have been proposed to acquire stable BVP signals for heart rate measurement. These techniques can be grouped into two categories: traditional signal processing methods and deep learning methods. 

Remote photoplethysmography (rPPG) has gained attention for its convenience and speed. Recent advancements in deep learning methods, including meta-learning, generative adversarial networks, and transformers, show significant potential in rPPG measurement \cite{ref3, ref4, ref5}. However, these methods often require subjects to face the camera with frontal facial orientations, leading to inaccuracies in face detection in real-world scenarios where subjects struggle to maintain stable postures. Non-frontal views can result in imprecise positioning of ROI pixels and undetected facial features, rendering ROI acquisition for signal extraction unfeasible \cite{ref7}. Furthermore, unique skin conditions and limited relevant databases for special patients pose challenges for current heart rate measurement methods, despite the crucial need for accurate heart rate acquisition for diagnosing and treating special patients. Additionally, utilizing face videos for heart rate measurement raises privacy concerns, particularly when collected without individuals' consent for biometric measurement purposes \cite{ref10}. The challenges are illustrated in Fig. 1.

\begin{figure*}[h]
	\centering
	\includegraphics[width=6.16in]{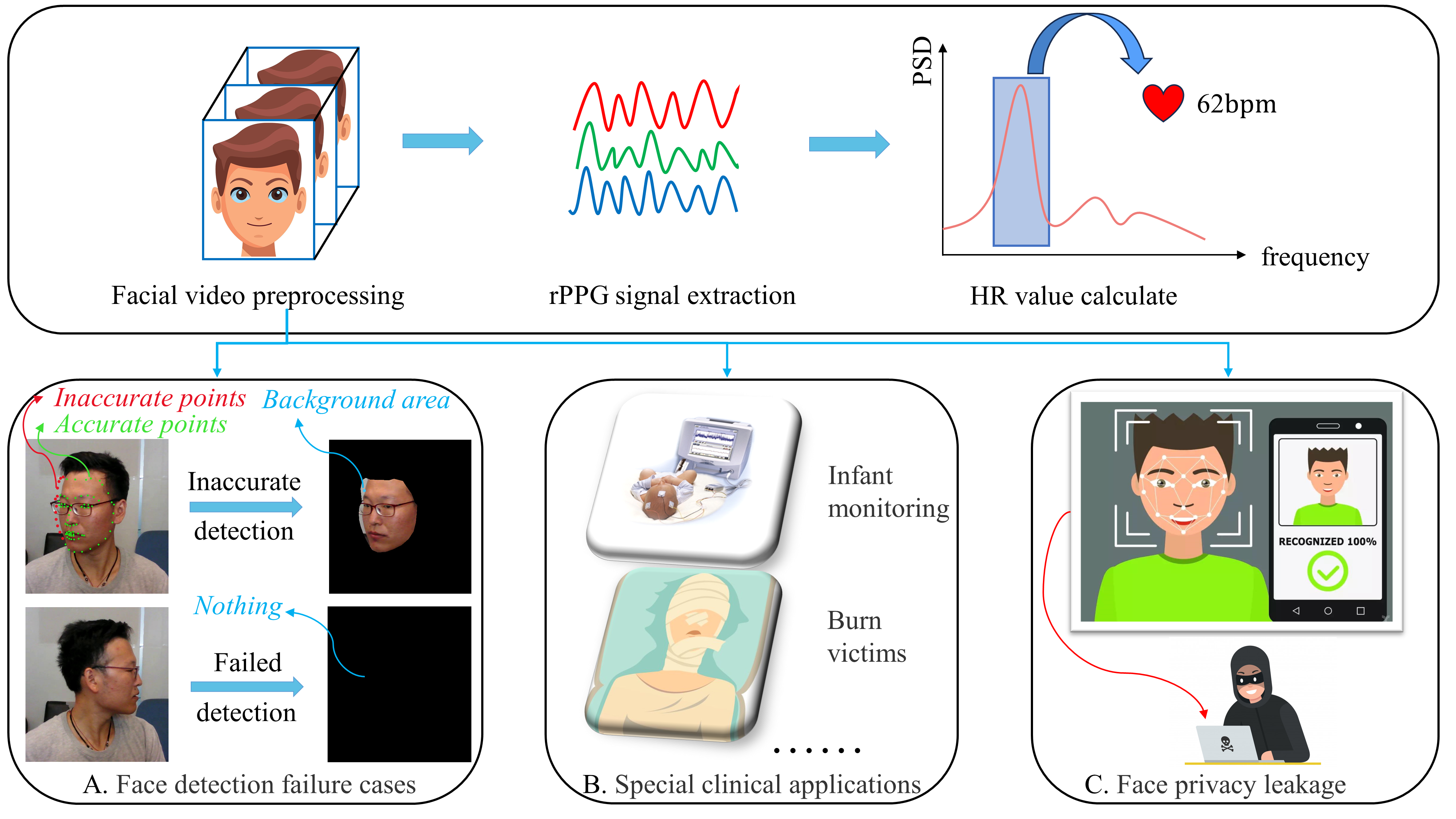}
	\caption{Challenges with existing heart rate measurement methods.}
	\label{fig1}
\end{figure*}

To address these issues, we design a robust motion-aware remote heart rate measurement framework, which can accurately identify skin ROI and be applied to body parts other than the face, making it applicable for special patients and addressing privacy protection concerns. Our framework contains three phases: (a) ROI extraction phase; (b) signal analysis phase; and (c) HR calculation phase.

In the ROI extraction phase, we introduced brain-inspired neural networks (continuous coupled neural network, CCNN) for robust skin tracking. CCNN encodes changing pixels into chaotic signals and static pixels into periodic ones, distinguishing skin from non-skin pixels. It excels in precise ROI extraction, overcoming head movement challenges. We have expanded this method to enhance privacy protection and extend to clinical applications like infant monitoring. In the signal analysis phase, we perform spatiotemporal feature analysis of optical flow signals within ROI videos. After extracting RGB signals, we filter the G-channel and conduct time-frequency analysis on temporal signals for heart rate calculation. Finally, mode values of heart rates at each pixel yield final measurements. This framework addresses inaccurate face detection, special patient applications, and privacy concerns, reducing dependency on extensive databases.

The main contributions of this work are summarized as follows:

\begin{itemize}
	\item We propose a novel remote heart rate measurement framework consisting of three phases: ROI extraction, signal analysis, and HR calculation. Each phase has a clear physical interpretation, ensuring high interpretability of the framework.
	\item To the best of our knowledge, the proposed ROI extraction method differs from all other methods relying on facial detection. In comparison, we introduce chaos theory to computer vision tasks, applying the video processing mechanism of the primary visual cortex for remote heart rate measurement.
	\item We show that this framework effectively extracts skin ROI, demonstrating robustness against motion and enabling reliable ROI extraction for subjects with head movements. It extends applicability to other body parts, ensuring applications for special patients and protecting individual privacy.
\end{itemize}  

\section{RELATED WORK}
Relevant prior work includes studies of heart rate measurement, traditional signal processing methods based on rPPG, and deep learning methods.

\subsection{Heart Rate Measurement}
HR measurement methods, encompassing contact and non-contact approaches, play vital roles in clinical and daily settings \cite{ref16}. While electrocardiography (ECG) is prevalent in clinical contexts, photoplethysmography (PPG) through wearable devices is common in daily life. However, both methods have limitations, such as discomfort and the need for skilled personnel and electrode application \cite{ref18,ref20,ref21}. In contrast, non-contact methods like remote photoplethysmography (rPPG), utilizing regular cameras, offer a solution by enabling contactless monitoring. Despite the advantages of rPPG under consistent ambient lighting, changes in lighting conditions and subject head movements pose challenges \cite{ref22}. These challenges, including dynamic lighting changes and subject movement complexities, have prompted the transition from traditional signal processing to deep learning methods \cite{ref23}.
\subsection{Traditional Signal Processing Methods Based on rPPG}
In the early stages of rPPG technology research, researchers emphasized blind source separation (BSS) techniques, like principal component analysis (PCA) \cite{ref24} and independent component analysis (ICA) \cite{ref25}, to extract clean BVP signals from mixed signals. However, this required extensive trial and advanced estimation techniques \cite{ref26}. Distinguishing signal distortions from pulse signals and periodic motion was challenging, hindering broader application. To address BSS limitations, De Haan \cite{ref28} introduced a standardized skin color vector and skin light reflection models across various color channels. Leveraging signal complementarity in different color spaces, he proposed the chrominance-based rPPG (CHROM) algorithm, mitigating subject head movement effects on pulse signal extraction. Wang \cite{ref29} expanded this method with the plane orthogonal to the skin (POS) algorithm, facilitating pulse signal extraction. Li \cite{ref30} addressed background illumination variation noise through illumination rectification, eliminated non-rigid motion caused by facial expressions like smiling, and improved HR measurement accuracy. Tulyakov \cite{ref31} introduced a self-adaptive matrix completion optimization framework, robustly estimating heart rate and dynamically selecting useful facial regions for measurement, overcoming noise from subject movements and facial expressions.

Traditional methods often require complex signal processing and rely on rPPG signal characteristics, skin reflection models, and prior assumptions. This makes them susceptible to variations in environmental illumination and individual differences, resulting in signal quality instability. While these methods have pioneered rPPG technology, the research focus has shifted towards deep learning methods due to these challenges \cite{ref23}.
\subsection{Deep Learning Methods}
Hsu \cite{ref32} introduced a deep learning framework for real-time pulse estimation using an RGB camera, achieving superior performance on public databases. Following this, research in rPPG technology shifted towards deep learning methods due to their strong modeling capabilities and in-depth exploration of algorithms. Lee \cite{ref3} argues that end-to-end learning methods use a global model during inference, which may not adapt well to unforeseen changes in real-world applications, affecting estimation performance. Therefore, he proposed a transductive meta-learner that adjusts weights using unlabeled samples during deployment, achieving fast adaptation to distribution changes. Niu \cite{ref33} constructed multi-scale spatiotemporal maps with face information and proposed a feature disentangling strategy to achieve robust physiological measurements. Lu \cite{ref4} points out that existing methods primarily focus on weak BVP signal enhancement without explicitly modeling noise, so he used Dual-GAN to jointly model BVP predictor and noise distribution, achieving robust remote physiological measurement. Yu \cite{ref5} suggests recent deep learning methods focus on mining subtle rPPG clues using CNNs with limited spatiotemporal receptive fields. He used temporal difference transformers to build a framework for long-range spatiotemporal perception, showing superior performance in intra- and cross-database experiments. Deep learning methods in rPPG technology leverage powerful modeling to mine the complex relationship between skin color changes and pulse signals, significantly improving measurement accuracy \cite{ref23}.

While deep learning methods have shown high accuracy, their lack of interpretability poses challenges in understanding how they learn or extract heart rate-related signals, especially when face detection is hindered by severe head movements \cite{ref36}. Despite their success on existing public databases, replicating such results in other databases or real-world applications is not guaranteed. Moreover, deep learning methods typically require large amounts of training data and long training times, demanding significant computing resources. Our method addresses these limitations by overcoming head movement tracking challenges, robustly extracting motion-unaffected skin areas, generating ROI synthetic videos, and performing time-frequency analysis on each pixel's time signal for accurate heart rate extraction. It offers good interpretability, strong generalization, and low computing resource requirements.

\begin{figure*}[h]
	\centering
	\includegraphics[width=6in]{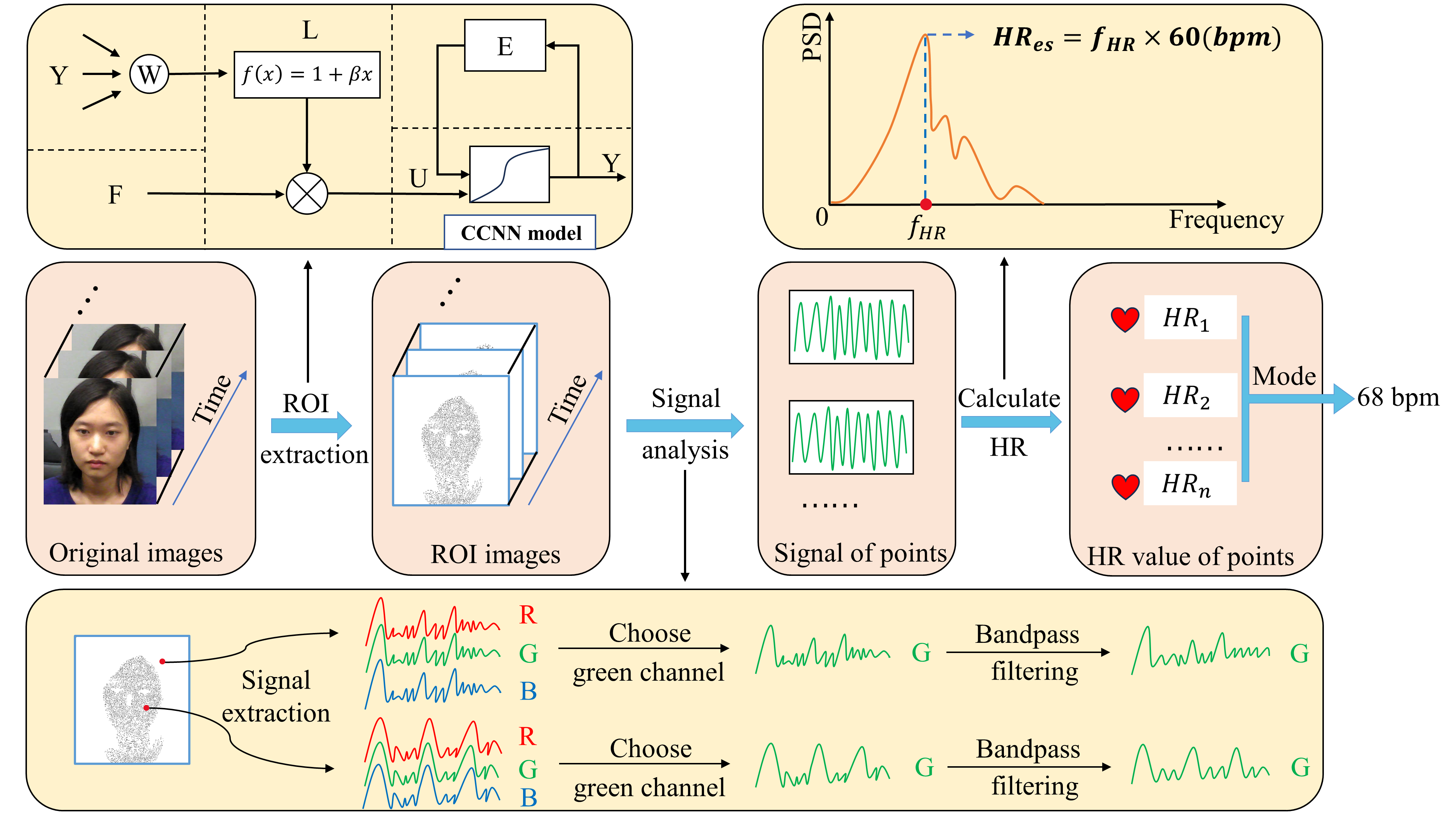}
	\caption{The framework of HR-RST.}
	\label{fig2}
\end{figure*}

\section{PROPOSED METHOD}
\subsection{Problem Statement and Challenges}
Extracting ROI is crucial in heart rate measurement tasks, often accomplished using established face detection algorithms like CLNF OpenFace2.2.0 \cite{ref41}, SeetaFace\footnote{https://github.com/seetaface/SeetaFaceEngine}, MTCNN face detector \cite{ref42}, etc. These algorithms offer high real-time performance by providing facial landmark points, which can guide ROI extraction either through the face frame or the coordinates of these points, as depicted in Fig. 1A. However, significant head movements can introduce inaccuracies in landmark positioning, leading to errors in recognizing facial edge pixels and potentially including environmental areas in the extracted ROI. Privacy concerns surrounding the use of face videos in remote heart rate measurement research have also become prominent, as they contain individually identifiable visual information. Additionally, recording face videos for database construction is costly and time-consuming, further complicating research efforts. Moreover, heart rate monitoring for special patients like infants and burn victims, characterized by complex movements and skin constraints, faces challenges due to limited relevant databases. Thus, exploring alternative body parts for heart rate measurement is imperative to address these existing issues \cite{ref45}.

\subsection{HR-RST}
To mitigate the impact of head movements and facilitate robust heart rate signal extraction, we conceptualize remote heart rate measurement as analyzing the spatiotemporal characteristics of optical flow signals in the video. The proposed framework, named HR-RST, is depicted in the flow chart shown in Fig. 2. It comprises three phases: ROI extraction, signal analysis, and HR calculation. The first phase aims to extract reliable skin ROI, the second phase involves signal analysis based on each point within the ROI synthetic video, and the third phase calculates the heart rate based on the signal analysis results from the second phase.

\subsection{ROI Extraction}
To address the issues outlined in Section \Roman{section}-A, this work utilizes CCNN for ROI extraction to tackle problems of face detection failure and to accurately extract ROI, thereby facilitating subsequent signal extraction tasks. The proposal of CCNN stems from recent advancements in neurodynamics. \cite{ref46} demonstrates that all pulse-coupled neural network (PCNN) models fail to explain the chaotic behavior of biological neurons stimulated by periodic signals. The analysis centers on the refractory period induced by neuron hyperpolarization, resulting in fluctuations in the dynamic threshold of the PCNN model. These fluctuations lead to differences in the dynamic characteristics of PCNN neurons compared to real neurons. The firing of action potentials in neuron clusters is assumed to follow a Gaussian process under specific membrane voltage. Under this assumption, the expression of the visual cortex neural network under the floating threshold is derived, called CCNN:

\begin{equation}
	\label{eq1}
	\left\{ {\begin{array}{*{20}{l}}
			{{F_{ij}}(n) = {e^{ - {\alpha _f}}}{F_{ij}}(n - 1) + {V_F}{M_{ijkl}}{Y_{kl}}(n - 1) + {S_{ij}}}\\
			{{L_{ij}}(n) = {e^{ - {\alpha _l}}}{L_{ij}}(n - 1) + {V_L}{W_{ijkl}}{Y_{kl}}(n - 1)}\\
			{{U_{ij}}(n) = {F_{ij}}(n)\left( {1 + \beta {L_{ij}}(n)} \right)}\\
			{{Y_{ij}}(n) = \frac{1}{{1 + {e^{ - \left( {{U_{ij}}(n) - {E_{ij}}(n)} \right)}}}}}\\
			{{E_{ij}}(n) = {e^{ - {\alpha _e}}}{E_{ij}}(n - 1) + {V_E}{Y_{ij}}(n - 1)}
	\end{array}} \right.
\end{equation}

CCNN consists of five main parts: dendritic coupling connection ${L_{ij}}(n)$, feedforward input ${F_{ij}}(n)$, modulation product ${U_{ij}}(n)$, dynamic threshold ${E_{ij}}(n)$, and continuous output ${Y_{ij}}(n)$. ${S_{ij}}$ is the external input received by the receptive field. Parameters ${\alpha _f}$, ${\alpha _l}$ and ${\alpha _e}$ are exponential decay factors that record the previous input state. ${V_F}$ and ${V_L}$ are weighting factors regulating the action potential of peripheral neurons. In addition, ${M_{ijkl}}$ and ${W_{ijkl}}$ are the weights of input and connecting synapses respectively, $\beta $ representing the connection strength, which directly determines the coupling connection ${L_{ij}}(n)$ in the modulation product ${U_{ij}}(n)$. CCNN structure is shown in Fig. 3. CCNN is a mean-field model, where the Heaviside function of PCNN output ${Y}$ is substituted with the sigmoid function. This substitution facilitates the training process, particularly for standard backpropagation. In fact, CCNN has demonstrated superior performance compared to the most advanced visual cortex model in image segmentation tasks \cite{ref51}.

\begin{figure}[!t]
	\centering
	\includegraphics[width=3.5in]{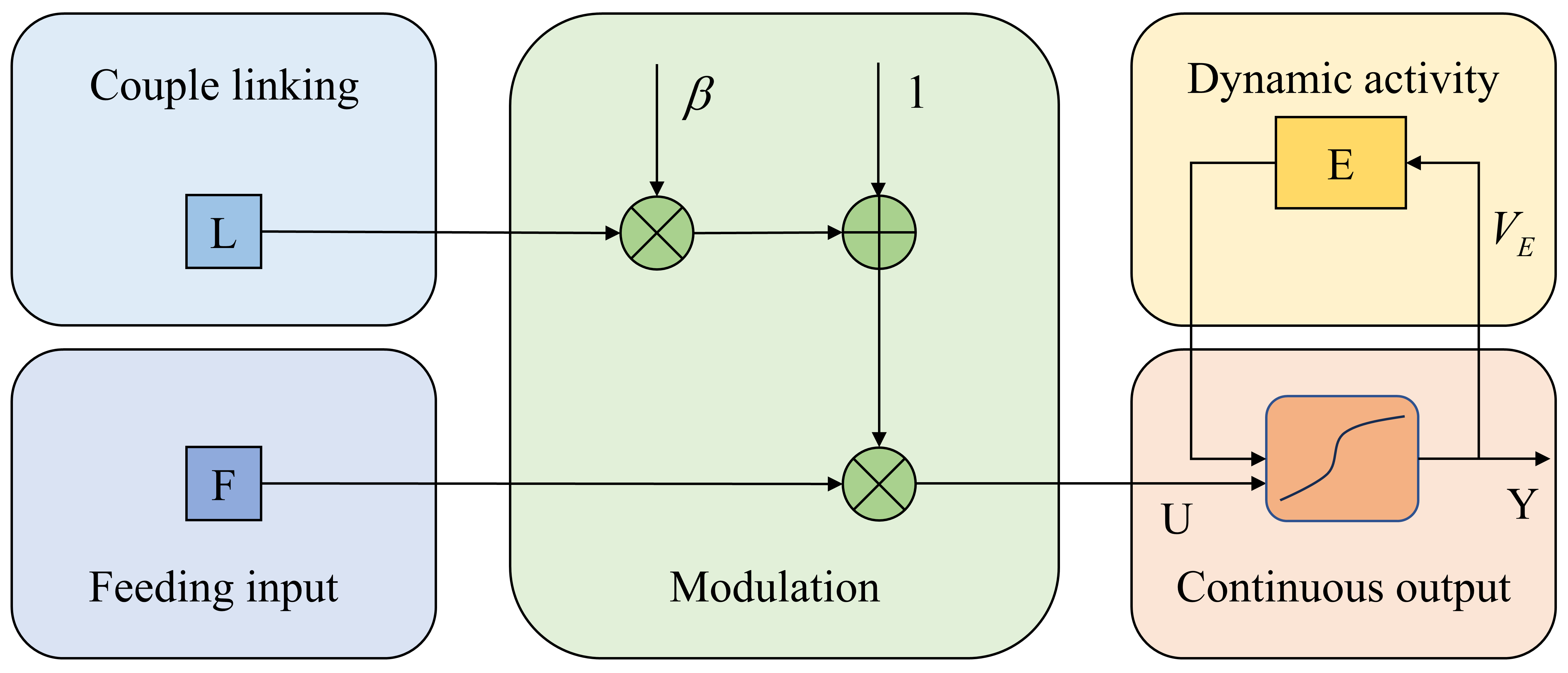}
	\caption{Structural diagram of CCNN.}
	\label{fig3}
\end{figure}

The characteristic of CCNN is to encode changing pixels into non-periodic chaotic signals and static pixels into periodic signals. Skin pixels exhibit changes through subtle color variations, while background regions remain relatively static. The recognition of moving target objects is performed by distinguishing the dynamic states corresponding to neuron clusters in the video. 

In the proposed framework, as depicted in Fig. 4, the ROI extraction phase comprises three steps. Firstly, we convert the color space of the original video from RGB to YIQ. Next, the CCNN model takes the time-varying I channel signal of each pixel in a continuous frame sequence of length three as input. The output is a time series of length three corresponding to the three-frame sequence, representing the pixel encoding result of the CCNN. Subsequently, the output signal from the CCNN neuron is processed by the continuous wavelet transform, and the frequency characteristics of the output signal are analyzed to distinguish between periodically changing pixels (skin area) and constant pixels (environment area), ultimately obtaining the ROI.

\begin{figure}[h]
	\centering
	\includegraphics[width=3.5in]{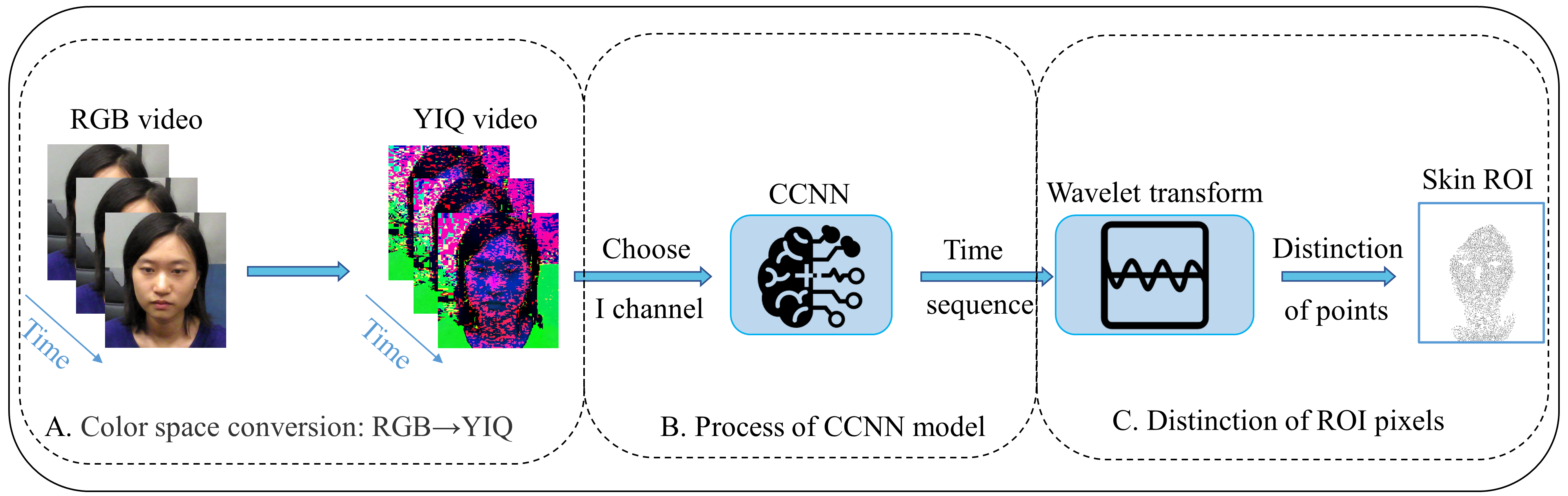}
	\caption{Flow chart of ROI extraction.}
	\label{fig4}
\end{figure}

In the continuous wavelet transform, we tried various continuous wavelets. Based on the experimental results, we selected the Gaussian wavelets. Gaussian wavelet function is a waveform function with local properties, and its definition is based on the Gaussian function. The Gaussian function is a continuously differentiable function with the following form:

\begin{equation}
	\label{eq2}
	\psi (t) = \frac{1}{{\sqrt {2\pi } \sigma }}{e^{ - \frac{{{t^2}}}{{2{\sigma ^2}}}}}
\end{equation}

\noindent where $t$ is the input signal, $\psi(\bullet)$ is the Gaussian function, and $\sigma$ is the standard deviation, which is used to control the width of the Gaussian function.

The Gaussian wavelet function is obtained by translation and scale transformation of the Gaussian function. Its definition is as follows:

\begin{equation}
	\label{eq3}
	{\psi _{a,b}}(t) = \frac{1}{{\sqrt a }}\psi \left( {\frac{{t - b}}{a}} \right) = \frac{1}{{\sqrt {2\pi a} \sigma }}{e^{ - \frac{{{{(t - b)}^2}}}{{2{a^2}{\sigma ^2}}}}}
\end{equation}

\noindent among them, $t$ is the input signal, ${\psi _{a,b}}(\bullet)$ is the Gaussian wavelet function, and $a$ is the scale parameter, which controls the scale of the wavelet function, and $b$ is the translation parameter, which controls the position of the wavelet function.

The Gaussian wavelet function operates in two main steps: decomposition and reconstruction. In the decomposition stage, the input signal or image undergoes wavelet transformation to generate wavelet coefficients across different scales and frequencies. This involves performing scale and frequency transformations on the coefficient matrix. Subsequently, in the reconstruction phase, the decomposed wavelet components are reconstructed back into the original signal or image. This is achieved by reversing the scale and frequency transformations and then applying inverse wavelet transformation. This process enables the input signal or image to be decomposed into wavelet components and subsequently reconstructed. Based on experimental findings, we opted for three scales.

According to the resulting ROI, the non-skin pixels in the first original frame among the three frames are set to be zero (a black pixel) while the RGB values of the remaining pixels are retained. Finally, the ROI synthetic video is obtained.

\subsection{Signal Time-frequency Analysis}
In the phase, we perform signal time-frequency analysis for all pixel points of ROI synthetic video. The phase consists of three steps, as shown in Fig. 5.

\begin{figure}[h]
	\centering
	\includegraphics[width=3.5in]{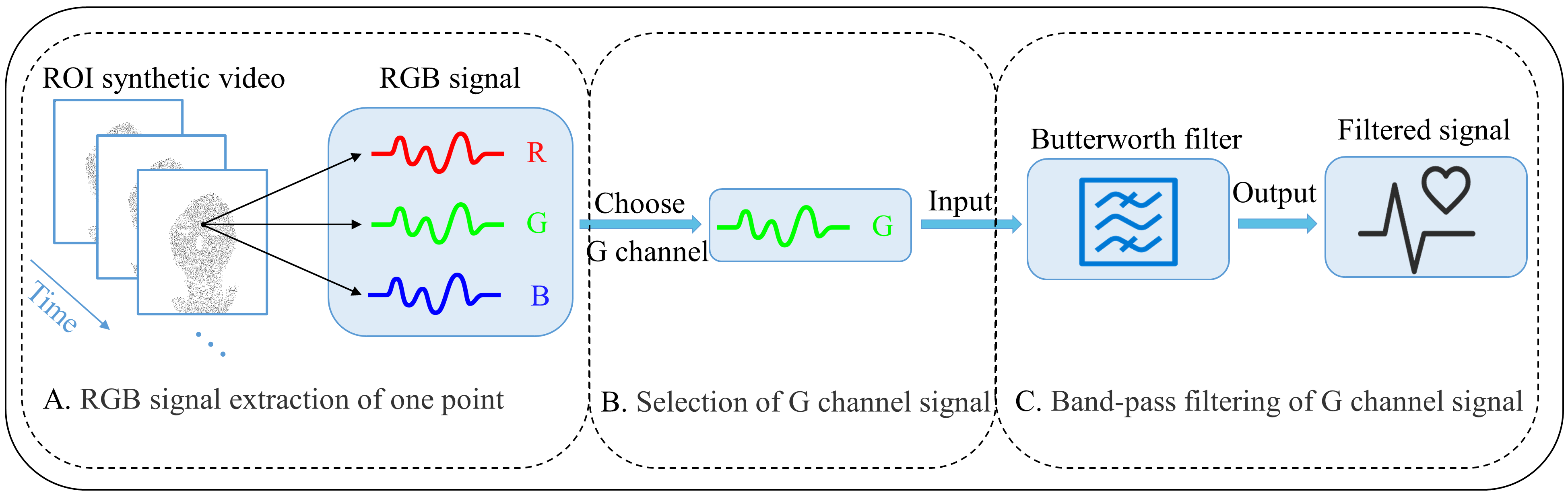}
	\caption{Flow chart of signal time-frequency analysis.}
	\label{fig5}
\end{figure}

We extract the RGB temporal signal from each pixel point of the ROI synthetic video, and then separate the green channel signal from the RGB temporal signal. Verkruysse \cite{ref56} highlighted that the green channel carries the most pertinent physiological information. This observation aligns with the fact that oxyhemoglobin exhibits greater absorption of green light compared to red light. Furthermore, green light's ability to penetrate deeper layers of the skin, making it suitable for blood vessel detection, adds to its relevance. Subsequently, we designed a third-order Butterworth band-pass filter to eliminate signals outside the desired frequency range. The band-pass frequency was defined as [0.7Hz, 2.5 Hz], corresponding to the typical heart rate range of [42bpm, 150bpm]. The filtered signal is used for heart rate measurement.

\subsection{Heart Rate Measurement}
Following the filtering process, we employed power spectral density (PSD) estimation to calculate the heart rate value. Similar to Li \cite{ref30}, we utilized the Welch method \cite{ref58}, which is an enhanced approach to classical power spectrum density estimation. This method overcomes the shortcomings of unsmooth power spectrum and large mean square error (MSE) obtained by periodogram spectrum estimation. The specific PSD process of each pixel point’s filtered signal is as follows:

The filtered signal of a pixel point is $x(n)$, $n$ denotes the time of the signal, and $n = 0,1, \cdots ,N - 1$, then $x(n)$ is divided with the time length of $N$ into $L$ segments, each segment has $M$ data (i.e., $N = L \times M$), and the data of the $i$-th segment is expressed as:

\begin{equation}
	\label{eq4}
	{x_i}(n) = x(n + iM - M),0 \le n \le M,1 \le i \le L
\end{equation} 

\noindent then apply the window function $w(n)$ to each data segment, $w(n)$ is the window function selected according to the task, and calculate the periodogram of each segment. The periodogram of the $i$-th segment is:

\begin{equation}
	\label{eq5}
	{I_i}(\omega ) = \frac{1}{U}{\left| {\mathop \sum \limits_{n = 0}^{M - 1} {x_i}(n)w(n){e^{ - j\frac{{2\pi }}{M}in}}} \right|^2},i = 1,2 \ldots ,M - 1
\end{equation} 

\noindent where $\omega $ is the signal frequency, $j = {( - 1)^{1/2}}$, and $U$ is called the normalization factor,

\begin{equation}
	\label{eq6}
	U = \frac{1}{M}\mathop \sum \limits_{n = 0}^{M - 1} {w^2}(n)
\end{equation} 

\noindent the periodogram of each segment is approximately regarded as uncorrelated, and the final power spectrum is estimated as:

\begin{equation}
	\label{eq7}
	{P_{xx}}\left( {{e^{j\omega }}} \right) = \frac{1}{L}\mathop \sum \limits_{i = 1}^L {I_i}(\omega )
\end{equation} 

\noindent where ${P_{xx}}(\bullet)$ is the final PSD estimation. The frequency with the maximum power spectral density is used as the frequency of heart rate. The frequency multiplied by 60 is the beats per minute (bpm):

\begin{equation}
	\label{eq8}
	H{R_{es}} = f[{\mathop{\rm argmax}\nolimits} ({P_{xx}}\left( {{e^{j\omega }}} \right))]{\kern 1pt} {\kern 1pt} {\kern 1pt} {\kern 1pt} {\kern 1pt} {\kern 1pt}  \times {\kern 1pt} {\kern 1pt} {\kern 1pt} {\kern 1pt} {\kern 1pt} 60{\rm{ }}({\rm{bpm}})
\end{equation} 

$H{R_{es}}$ is the estimated heart rate value, and $f(\bullet)$ is the heart rate frequency.


Following heart rate computation from all ROI synthetic video points, histogram analysis in Fig. 6A illustrates their distribution. Except for values near zero, representing environmental background, most points cluster around the ground truth heart rate of 69 bpm. Additionally, Fig. 6B presents a thermographic representation of heart rate distribution, with points related to the environment displaying higher errors compared to those aligned with true values, mainly within the skin region, showing lower errors. Consequently, we adopt the mode value among these values as the final output result.


\begin{figure}[h]
	\centering
	\includegraphics[width=3.5in]{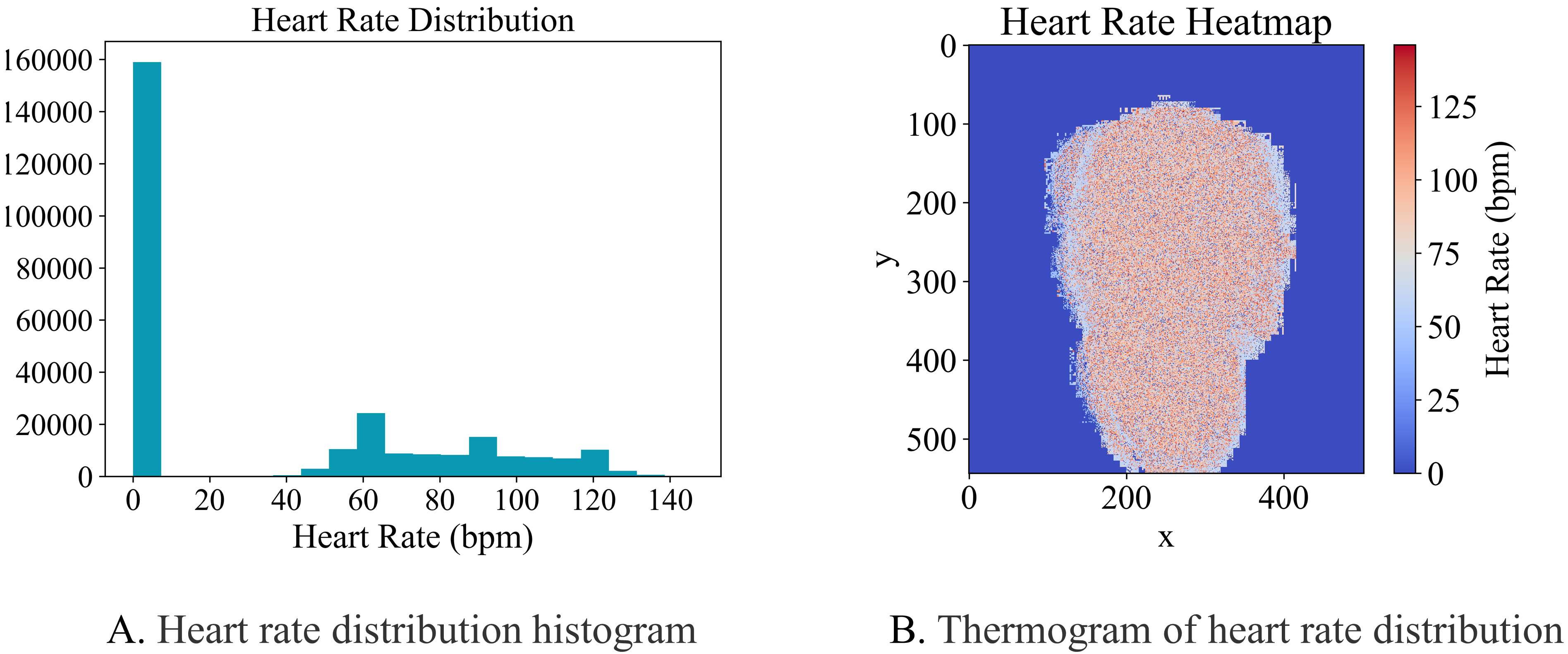}
	\caption{Statistical analysis of heart rate distribution.}
	\label{fig6}
\end{figure}

\section{EXPERIMENT}
To evaluate the effectiveness of HR-RST, we performed two experiments: (a) experiments involving data collected from various body parts collected by ourselves, and (b) heart rate measurement experiments using the VIPL-HR database.

\subsection{Databases and evaluation indicators}
VIPL-HR \cite{ref59} is a comprehensive database established by the VIPL Research Group at the Institute of Computing Technology, Chinese Academy of Sciences, comprising data from 107 subjects across 9 distinct scenarios. These scenarios encompass varied conditions such as static scenes, lighting changes, and head movements, resulting in 2378 RGB videos captured using 4 different devices. Ground truth heart rate measurements synchronized with the recordings are included. To ensure fair comparisons with prior methods, we calculated the average heart rate of each video and assessed measurement error alongside statistical metrics like standard deviation (SD), mean absolute error (MAE), and root mean square error (RMSE). For experiments involving non-facial body parts, we conducted data collection due to the absence of relevant research and public databases.

\subsection{Experiments on Other Body Parts}
We utilized CCNN to process every three consecutive frames of the original video frames. Fig. 7 depicts the processing of two points in the initial three frames: one on the palm and the other in the background. This involved extracting the I-channel signal, obtaining a time series via CCNN, and performing wavelet transform on the time series. Discrimination of ROI points relied on the sum of real parts of the wavelet transform results: if the sum exceeded zero, the pixel was classified as an ROI point; otherwise, it was considered non-ROI. For the first frame, non-ROI pixel values were set to grayscale zero, while RGB values remained for the rest. This process was applied to all frames, yielding the final ROI synthetic video. In addressing subject privacy, we conducted remote heart rate experiments on six non-facial body parts, reducing individual identifiable information. These parts included the palm, back of the hand, forearm, upper arm, back, and sole. Fig. 8 displays the results of ROI extraction for these body parts.


\begin{figure}[h]
	\centering
	\includegraphics[width=3.5in]{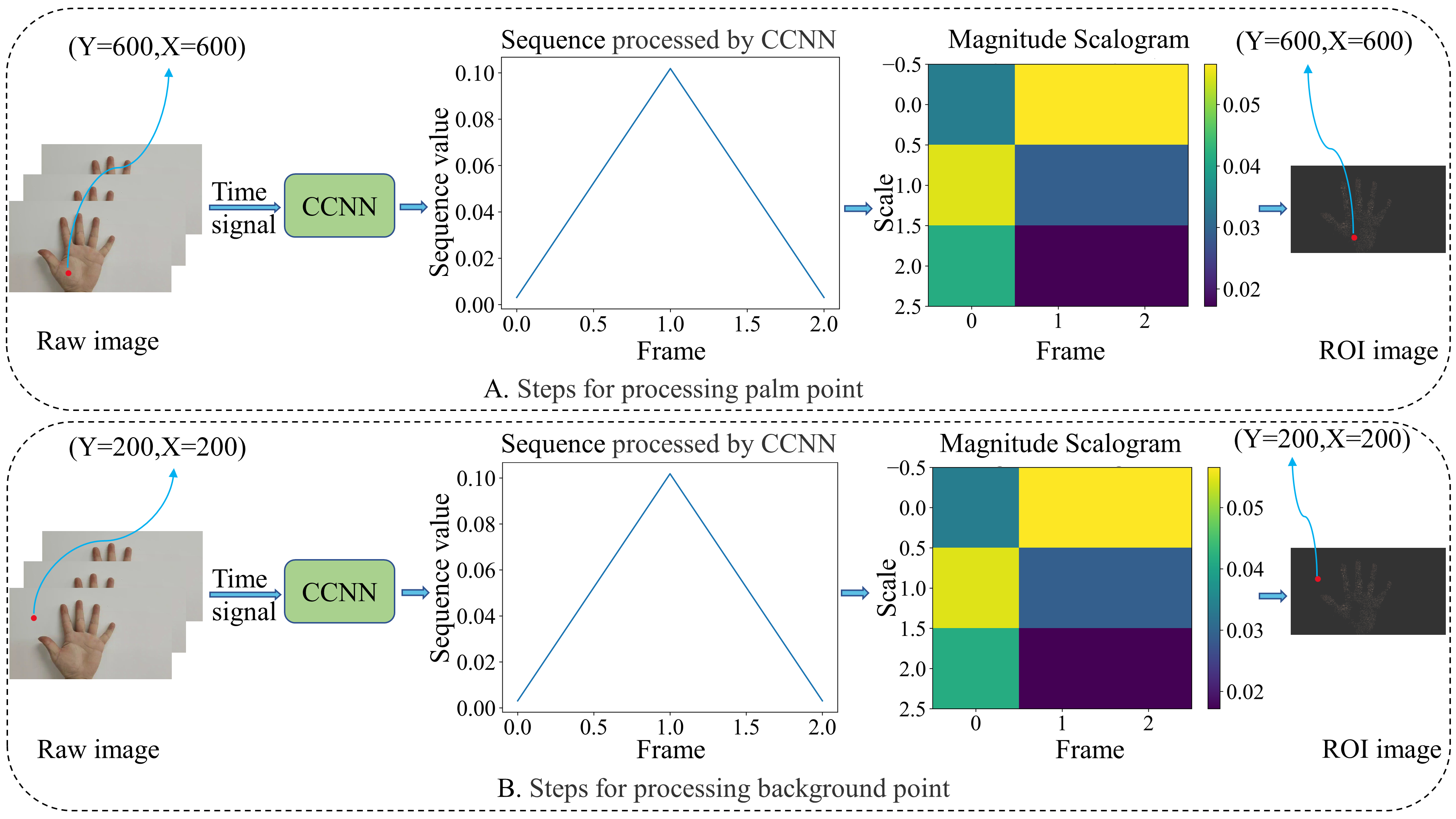}
	\caption{The discrimination process of HR-RST for ROI points on the palm and background points.}
	\label{fig7}
\end{figure}


\begin{figure}[h]
	\centering
	\includegraphics[width=3.5in]{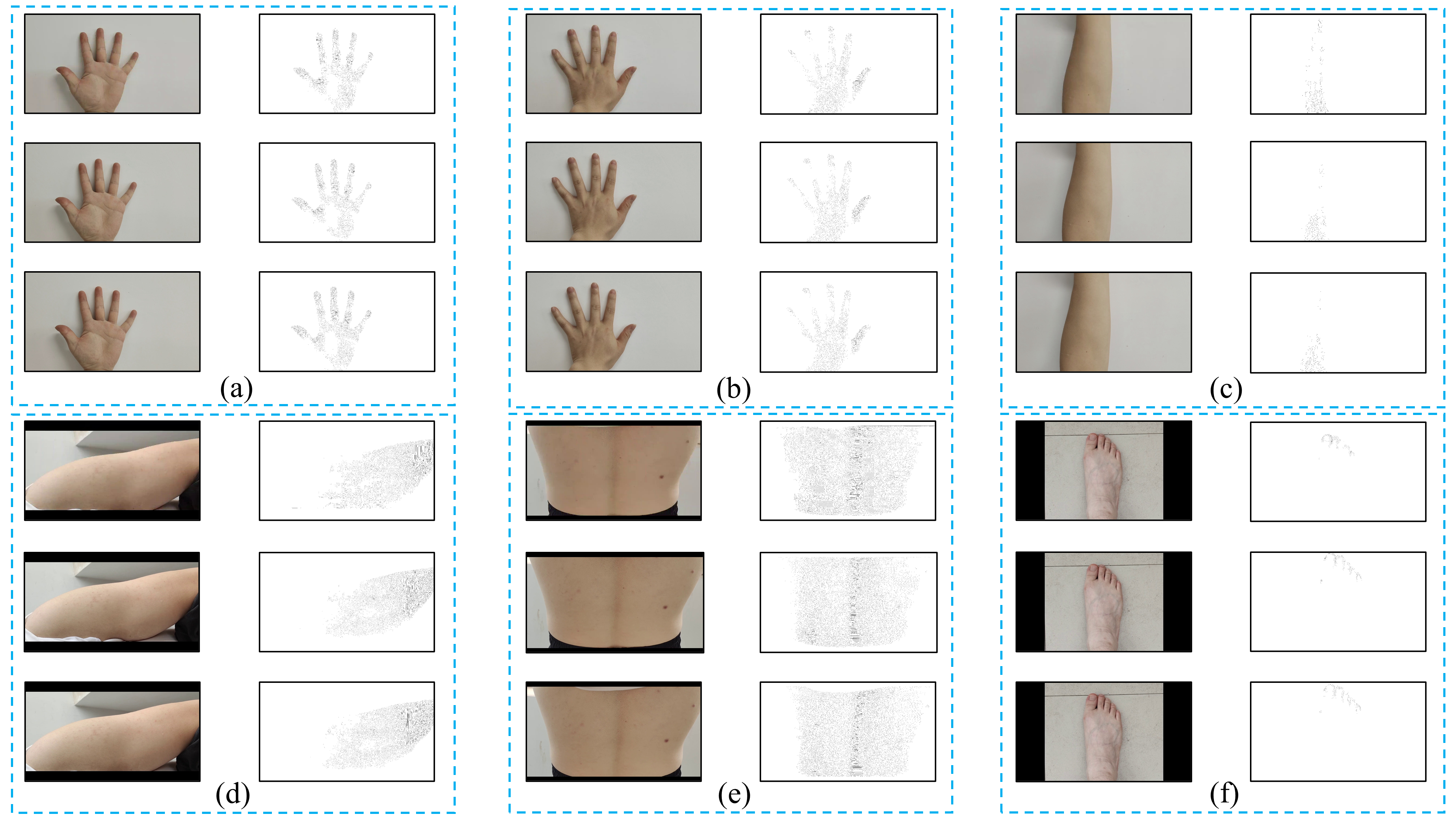}
	\caption{ROI extraction of other body parts, including (a) palm and its extraction results; (b) back of hand and its extraction results; (c) forearm and its extraction results; (d) upper arm and its extraction results; (e) back and its extraction results; (f) sole and its extraction results.}
	\label{fig8}
\end{figure}

In the above results, except for only a small amount of skin area extracted from the forearm and sole, relatively accurate ROIs were obtained in the other four body parts. These images will be combined into ROI synthetic videos for heart rate measurement. We argue that CCNN shows immense for potential ROI extraction capabilities. Table~\ref{tab1} shows the heart rate measurement results obtained using these body parts. Among them, $H{R_{gt}}$ is the ground truth value of the heart rate, $H{R_{es}}$ is the heart rate estimate value, and $H{R_{er}}$ is the heart rate error value, that is, $H{R_{er}} = H{R_{es}} - H{R_{gt}}$.

\begin{table}[h]
	\centering
	\caption{Heart rate measurement results of six body parts ($H{R_{gt}}$=\textnormal{65 bpm})}
	\label{tab1}
	\begin{tabularx}{\columnwidth}{l *{6}{>{\centering\arraybackslash}X}}
		\toprule
		\textbf{Body site} & \textbf{Palm} & \textbf{Back of hand} & \textbf{Forearm} & \textbf{Upper arm} & \textbf{Back} & \textbf{Sole} \\
		\midrule
		$H{R_{es}}$ (bpm)  & 43  & 30  & 43  & 43  & 43  & 43  \\
		$H{R_{er}}$ (bpm)  & -22  & -35 & -22 & -22 & -22 & -22 \\
		\bottomrule
	\end{tabularx}
\end{table}

HR-RST extracts heart rate from unconventional non-facial areas like the back of the palm, showcasing the potential of various body parts besides the face for physiological measurement. This not only addresses the increasing concern for privacy protection but also lays the groundwork for more advanced physiological measurements in special clinical scenarios, such as infant monitoring and care for burn victims.

\subsection{Evaluation Experiment on VIPL-HR}
We fed each set of three original video frames into CCNN for processing. Fig. 9 showcases the processing of two points on the forehead and background in the initial three frames, involving I-channel signal extraction, time sequence acquisition from CCNN, and wavelet transform result depiction. The ROI frame was determined based on the sum of real parts of the wavelet transform: if the sum exceeded zero, it was considered an ROI point; otherwise, a non-ROI point. Non-ROI pixel values in the first frame were set to zero, while RGB values of the remaining pixels were retained. This process was applied to all frames, yielding the final ROI synthetic video. Experimental results in Fig. 10 demonstrate that compared to ROI extraction by the face detector, CCNN-extracted ROIs comprise only facial and neck skin, excluding environmental areas that may affect pulse signal extraction. Moreover, the model effectively tracks subject head movements and accurately locates skin edge pixels, segregating accessories like glasses and necklaces while capturing facial expressions like smiling with an open mouth. Crucially, the proposed method overcomes previous limitations in recognizing faces during head tilts due to pitch and yaw rotation.


\begin{figure}[h]
	\centering
	\includegraphics[width=3.5in]{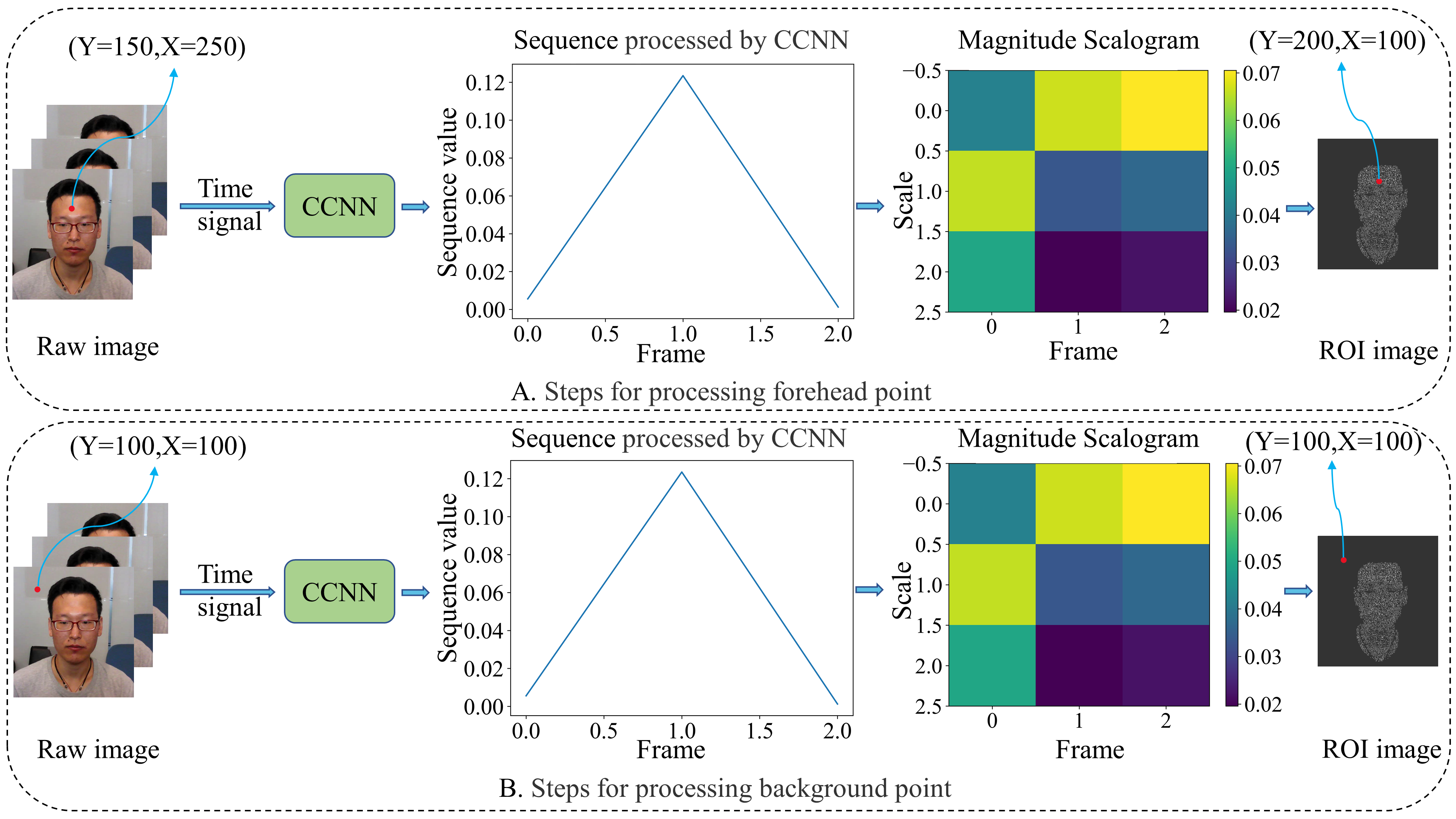}
	\caption{The discrimination process of HR-RST for ROI points on the forehead and background points.}
	\label{fig9}
\end{figure}


\begin{figure}[h]
	\centering
	\includegraphics[width=3.5in]{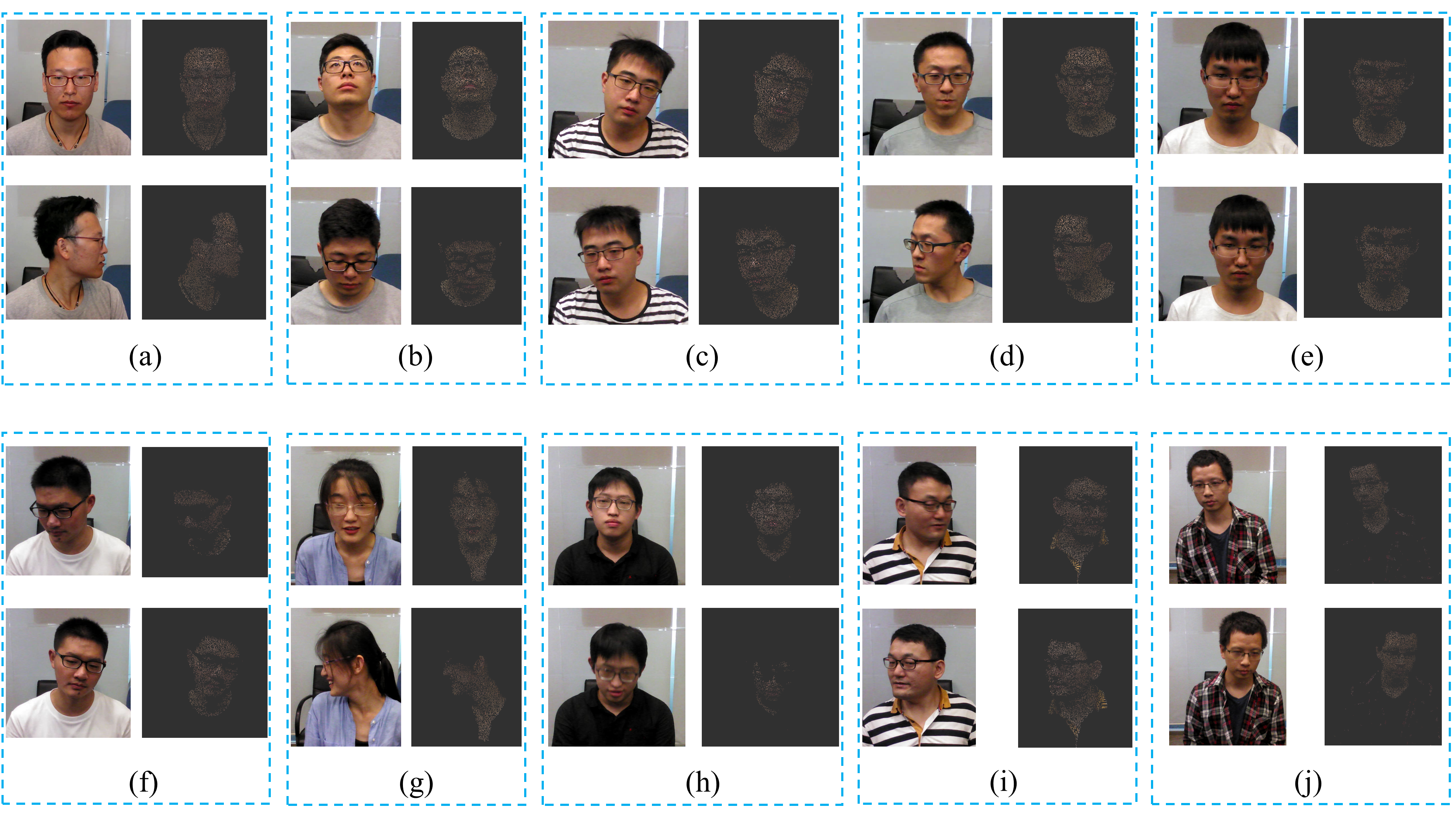}
	\caption{Example of the result map of HR-RST recognition of skin area, including ten subjects, (a) p1 and its extraction results; (b) p2 and its extraction results; (c) p3 and its extraction results; (d) p4 and its extraction results; (e) p5 and its extraction results; (f) p6 and its extraction results; (g) p7 and its extraction results; (h) p8 and its extraction results; (i) p9 and its extraction results; (j) p10 and its extraction results.}
	\label{fig10}
\end{figure}

In the VIPL-HR database, we conducted experiments using all RGB videos from ten subjects labeled as p1 to p10. To preserve physiological information without loss due to video compression, we utilized the original videos from the database. HR-RST obviates the need for complex image pre-processing and training procedures, as it directly computes heart rates following the acquisition of the ROI synthetic video. The results of these measurements for the ten samples are presented in Table~\ref{tab2}.

\begin{table}[!t]
	\centering
	\caption{HR estimation results of HR-RST on VIPL-HR database}
	\label{tab2}
	\begin{tabularx}{\columnwidth}{l *{6}{>{\centering\arraybackslash}X}}
		\toprule
		\textbf{Sample} & \textbf{SD↓} & \textbf{MAE↓} & \textbf{RMSE↓}  \\
		\midrule
		\textbf{p1}  & 13.29  & 12.41  & 13.74  \\
		\textbf{p2}  & 12.69  & 26.41  & 29.18  \\
		\textbf{p3}  & 10.53  & 16.22  & 18.56  \\
		\textbf{p4}  & 17.63  & 15.05  & 19.54  \\
		\textbf{p5}  & 15.99  & 27.14  & 29.37  \\
		\textbf{p6}  & 13.85  & 13.00  & 16.17  \\
		\textbf{p7}  & 6.37  & 30.17  & 30.81  \\
		\textbf{p8}  & 7.38  & 32.17  & 32.97  \\
		\textbf{p9}  & 13.03  & 24.48  & 27.60  \\
		\textbf{p10}  & 14.56  & 19.22  & 21.36  \\
		\bottomrule
	\end{tabularx}
\end{table}

From the experimental results, it's evident that HR-RST consistently produces comparable metrics across ten samples, indicating the robustness of our method to different sample variations. Based on this, we contend that our framework can overcome the challenges posed by complex scenarios in the VIPL-HR database and yield reliable remote heart rate measurement results.

\section{CONCLUSION}
Heart rate measurement using commonly available cameras is a welcome technological advance. Although the popular deep learning methods have achieved great success in the field, they encounter problems in real-world applications such as the scarcity of large databases and the difficulty of high cost of collecting data. Different from existing methods, this work applies chaos theory to computer vision tasks for the first time, we have proposed a robust motion-aware framework. It overcomes the problem that other face detection methods cannot detect faces when the head moves significantly, and it has strong interpretability. Concluding from the experimental results, there is a gap between the indicators of the framework and the best deep learning methods. Considering the dependence of deep learning methods on data, the results obtained in the work are promising. We anticipate the work to be a starting point for more sophisticated physiological measurement in special clinical scenarios such as infant monitoring and burn victims.

\section*{Acknowledgment}
This work is supported by the Natural Science Foundation of Gansu Province (Grants 21JR7RA510, 21JR7RA345, 22JR5RA543).

Some experiments are supported by the Supercomputing
Center of Lanzhou University.

Conflict of Interest: The authors declare that they have no conflicts of interest.

\bibliographystyle{IEEEtran}
\bibliography{mybib}

\end{document}